\documentclass[journal]{IEEEtran}
\IEEEoverridecommandlockouts
\usepackage{comment}
 
\usepackage{amsmath,amssymb,amsfonts}
\usepackage{algorithmic}
\usepackage{graphicx}
\usepackage{xcolor}
\usepackage{multirow}
\usepackage[utf8]{inputenc}
\usepackage[english]{babel}
\usepackage[style=numeric,sorting=none]{biblatex}
\usepackage{csquotes}
\usepackage{biblatex}
\bibliography{sample.bib}
\AtBeginBibliography{\small}
\begin{document}
\title{Unleashing the Power of Unlabeled Data: A Self-supervised Learning Framework for Cyber Attack Detection in Smart Grids}

\author
{\IEEEauthorblockN{Hanyu Zeng\IEEEauthorrefmark{1}\IEEEauthorrefmark{5}, Pengfei Zhou\IEEEauthorrefmark{5}, Xin Lou\IEEEauthorrefmark{1}\IEEEauthorrefmark{4},   Zhen Wei Ng\IEEEauthorrefmark{1},
David K.Y. Yau\IEEEauthorrefmark{1}\IEEEauthorrefmark{2},
Marianne Winslett\IEEEauthorrefmark{1}\IEEEauthorrefmark{3}}
	\IEEEauthorblockA{\\
		\IEEEauthorrefmark{1}Illinois Advanced Research Center Singapore
		\IEEEauthorrefmark{2}Singapore University of Technology and Design\\
		\IEEEauthorrefmark{3}University of Illinois at Urbana-Champaign, USA
		\IEEEauthorrefmark{4}Singapore Institute of Technology\\
		\IEEEauthorrefmark{5}University of Pittsburgh, USA
		% \IEEEauthorrefmark{6}National University of Singapore
	}
}
\IEEEoverridecommandlockouts

% \IEEEpubid{\makebox[\columnwidth]{978-1-6654-3254-2/22/\$31.00~\copyright2022 IEEE \hfill}\hspace{\columnsep}\makebox[\columnwidth]{ }}

\maketitle
\IEEEpubidadjcol

\begin{abstract}
Modern power grids are undergoing significant changes driven by information and communication technologies (ICTs), and evolving into smart grids with higher efficiency and lower operation cost. Using ICTs, however, comes with an inevitable side effect that makes the power system more vulnerable to cyber attacks. In this paper, we propose a self-supervised learning-based framework to detect and identify various types of cyber attacks. Different from existing approaches, the proposed framework does not rely on large amounts of well-curated labeled data but makes use of the massive unlabeled data in the wild which are easily accessible. Specifically, the proposed framework adopts the BERT model from the natural language processing domain and learns generalizable and effective representations from the unlabeled sensing data, which capture the distinctive patterns of different attacks. Using the learned representations, together with a very small amount of labeled data, we can train a task-specific classifier to detect various types of cyber attacks. 
Meanwhile, real-world training datasets are usually imbalanced, i.e., there are only a limited number of data samples containing attacks. In order to cope with such data imbalance, we propose a new loss function, separate mean error (SME), which pays equal attention to the large and small categories to better train the model.
Experiment results in a 5-area power grid system with 37 buses demonstrate the superior performance of our framework over existing approaches, especially when a very limited portion of labeled data are available, e.g., as low as 0.002\%. We believe such a framework can be easily adopted to detect a variety of cyber attacks in other power grid scenarios.
\end{abstract}

\section{Introduction}
% Recently, Cyber-Physical System(CPS) has been attracting more attention due to the explosion of artificial intelligence. Unconsciously, the CPSes have played an indispensable role in people's daily life. For example, the product lines in factories, the smart power grid around people, and even the electronic products often used in life are all inextricably linked to Cyber-physical systems. However, due to the increasing importance of CPS, anomaly detection in those systems become a main task for researchers. 
% The power system is one of the most familiar cyber-physical systems for people, the electricity used in daily life is all translated by the power grid. Although it plays an irreplaceable role, the system is still vulnerable to some cyber attacks.

% Smart grid has been equipped with networks of sensors and generators allowing two-way communication within the system with information and communication technologies (ICTs), which can help the operators manage a larger scale area of power distribution~\cite{sakhnini2019smart}. This feature, however, also makes the power system more vulnerable to cyber attacks, e.g., false data injection attack (FDIA)~\cite{yanSFDIA} and time delay attack (TDA)~\cite{sargolzaeiTDA}. The purpose of cyber attacks is mainly to cause drastic frequency passivity and finally crash the whole system~\cite{lou20}. 
The smart grid's use of information and communication technologies (ICTs) to allow two-way communication between sensors and generators has improved the ability of operators to manage a larger scale area of power distribution~\cite{sakhnini2019smart}. However, this feature also makes the power system more vulnerable to cyber attacks, such as false data injection attacks (FDIA)~\cite{yanSFDIA} and time delay attacks (TDA)~\cite{sargolzaeiTDA}. The purpose of cyber attacks is mainly to cause drastic frequency instability and ultimately bring down the entire system~\cite{lou20}.

The automatic generation control (AGC) system is one of the most important systems in power grids, but it is also vulnerable to cyber attacks~\cite{lou20,Tan16}. AGC adjusts the output of generators to keep the system frequency within a safe range. A breach of this safe range due to frequency excursion caused by cyber attacks can cause damage to the system~\cite{yanSFDIA,lou20}.
In this work, we consider a practical scenario where AGC is distributed over a large area, with networked sensors collecting sensing data such as the system frequency and power export. In a real-world system, we do not know when or where attacks will occur, and the massive amount of sensing data collected in the wild is unlabeled. Manually collecting and labeling sensing data for different cyber attacks is expensive and time-consuming. As a result, it is challenging to use the limited amount of labeled data to develop effective models for detecting different cyber attacks in real-time.
A number of mechanisms have been proposed to detect and identify cyber attacks in power systems~\cite{Ganesh9,louxin12,detFDI25,FDILSTM26,anKNN,sakhniniUSL}. Some of these mechanisms use supervised learning models, which require a large number of labeled data to achieve accurate attack detection. However, this is not scalable for real-world systems. Unsupervised learning methods, including traditional machine learning (ML) methods such as K-nearest neighbors (KNN)~\cite{anKNN} and one-class support vector machines (OCSVM)~\cite{uddinOCSVM} and deep learning (DL) methods such as stacked denoising autoencoders (SDAE)~\cite{HuSDAE} and weighted convolutional autoencoders (WPD)~\cite{yangWPD}, have been proposed to detect anomalies in power grids using unlabeled data.
In practice, different cyber attacks have completely different means to disrupt the power system, which indicates that different countermeasures need to be taken against them. Therefore, it is not sufficient to simply detect anomalies. Recent works~\cite{HuSDAE,yangWPD,dairi2023semi} have leveraged unsupervised learning to detect specific types of attacks, such as false data injection attacks (FDIA)~\cite{HuSDAE} and time delay attacks (TDA)~\cite{yangWPD}. However, these methods only target a single specific type of attack. ~\cite{dairi2023semi} trained various versions of models to do detection on different attacks.
Furthermore, the effects of different attacks can vary across the spatial domain. This means that the impact of an attack on one region of the power grid may not be the same as its impact on another region. This discrepancy can have an impact on the detection performance of the attack. However, none of the current studies have taken this factor into account.

In this paper, we aim to take a step further towards making multi-type cyber attack detection and classification with the knowledge learned from massive unlabeled sensing data and propose PowerBERT, a BERT-like \cite{Devlin4} self-supervised learning model to deal with the sensing data in smart grids for cyber attack detection. The proposed PowerBERT learns effective and generalizable representations from massive unlabeled sensing data collected in the wild. Once the representations have been learned, together with a small amount of labeled data for the targeted types of attacks, we can easily train a task-specific classifier to detect various types of attacks. In addition, we explore the impact of spatial effects on the performance of our model, as previously discussed, and have also explored approaches to self-supervised learning to mitigate the negative impacts of imbalanced datasets.
% {\color{red}{Besides, we investigate the impact of spatial effects on the performance of our model.}}
%{\color{red}{Additionally, those two types of attack have both timely and spatial impacts on the system. We compare the performance of the model at different time durations and in different regions to analyze the range of impact of the two attacks in time and spatial domains in the system. With this information, we can determine a more precise scope of examination in future attack detection and analysis.}}
% And the representations will be used for following downstream tasks.
% In our paper, we focus on the time delay attack and false data injection attack. The PowerBERT model is a self-supervised learning model which means it learns latent features from unlabeled data and then classifies and detects attacks based on the learned feature by using a few labeled data. We test our model on the area control errors(ACEs) data of an AGC system and the results demonstrate our model outperforms other competitors on a small amount of labeled dataset.

The BERT was originally designed for natural language processing (NLP) and lacks the methodology to deal with sensing data in power grids where the data distributions are different and require in-depth investigation. 
%Previously, researchers made some improvements on it and used it in learning IMU data \cite{xu2021limu} \textcolor{red}{(one new sentence)}
Inspired by the observation that cyber attacks usually cause both temporal and spatial signal variation across the power girds, in this paper, we propose to learn effective representations with the sensing data collected from neighboring areas in a region. 
% We segment the time-series sensing data into data partitions and each partition corresponds to an \textit{event}. 
A series of data clips extracted by sliding window are fed into PowerBERT to learn effective representations that can capture the spatial-temporal signal fluctuation patterns caused by cyber attacks, and the patterns caused by different attacks are distinctive.
% From our observation, the impact of the attack in the system is not immediately but generally, and the impact is also spatially transmissive. So instead of feeding the data individually, we combine some spatially and temporally adjacent data and input them together, and we regard it as an event and the process as an event mechanism.

% For the reason that the attacked data are more complex to recover\cite{zong18}, we suppose the information involves inside the reconstruction errors are also effective in anomaly detection task.

%From observation, we found that most of the datasets in the real systems are imbalanced, especially in the security field, because attack by cyber-attacks is occasional events. The approach in~\cite{zengPowerBERT} works well in attack identification in smart grids when the unlabeled dataset is balanced, but cannot handle the imbalanced data problem in self-supervised learning. 
In the meantime, as previously stated, we continue to enhance the performance of our self-supervised learning model by addressing the challenges posed by imbalanced datasets. In real-world applications, training datasets are often imbalanced, i.e., there are only a limited number of data samples containing attacks. This imbalance can cause self-supervised learning models to learn better features of the larger categories and overlook those of the smaller categories, leading to larger training losses for those smaller categories. However, because these larger loss categories make up a small portion of the dataset, these large training losses may be easily overlooked by most of the commonly used loss functions such as mean absolute error (MAE) and mean square error (MSE). The approach in~\cite{zengPowerBERT} works well in attack identification in smart grids when the unlabeled dataset is balanced, but cannot handle the imbalanced data in practice. To address this challenge, we propose a new loss function called separate mean error (SME), which can balance  the impact caused by the imbalanced dataset.  Different from those commonly used loss functions, SME tries to pay equal attention to the large and small categories, by separately calculating the loss for both of them. Thus, it can ensure that the training loss is not only small but also balanced, leading to improved model performance in practical scenarios. And due to the characteristics of SME, it can be well adapted to data sets of different imbalance degrees.

Utilizing this novel loss function, the model is now able to assimilate insights from the reconstruction error throughout the auto-encoder training process. As a result, we opt to forego the Gaussian Mixture Model~\cite{GMM} approach for extracting latent features from within the reconstruction errors. This alteration obviates the necessity of invoking the decoder component during model inference, leading to a substantial reduction in computational overhead.

% % Compared with the magnitude, the distributions of reconstruction errors are considered to be more informative.
% In PowerBERT, the distributions of the reconstruction errors are also utilized as features in the downstream classification as well as representations.

% As for the application of those representations, we used them for cyber attack detection and classification. In our paper, we mainly focus on FDIA and TDA two kinds of attack, and we trained random forest classifier with a small amount of labeled data.
As for the dataset, we have collected a new dataset from a larger multi-area smart grid system than the previous paper~\cite{zengPowerBERT}, which involved 5 areas, in order to explore the spatial effects of those attacks in a more complex system. This dataset contains 11826 traces, including normal, FDIA, and TDA data. Additionally, this new dataset helps to demonstrate the generalizability of our model to different systems and attack rates.

We show the effectiveness of the learned representations for detecting the FDIA and TDA with a random forest classifier. By leveraging a very small amount of labeled data (i.e., 0.002\%), the proposed model can achieve 93.8\% and 87.2\% detection accuracy for FDIA and TDA, respectively. Using 0.002\%$\sim\, $0.02\% labeling rate, PowerBERT-based method outperforms existing models at least by 20.0\% to 2.5\% in terms of F1-score on average.

The main contributions of this paper are as follows:
\begin{itemize}
    \item We propose PowerBERT, an auto-encoder inspired by BERT, designed to acquire comprehensive and efficient representations using abundant unlabeled sensor data from adjacent zones within smart grids. Our approach involves partitioning the time-series sensor data into various window sizes and subsequently refining the optimal setup for detecting diverse forms of attacks within the AGC control of power grids.
    % We proposed a self-supervised auto-encoder to learn the latent representation from the unlabeled data in the power system. And based on our model, we propose the event mechanism to train a better feature extraction model. % and the classifier model can be trained by a small number of labeled samples, which significantly reduced the cost of obtaining labeled data in supervised training.
    \item We train a random forest classifier based on the learned representations with a small amount of labels for FDIA and TDA. The classifier can be easily adopted to detect other types of attacks with corresponding labeled data.
    % We apply representations to downstream tasks. In our work, the classifier for FDIA and TDA is trained based on a small amount of labeled data. By using the extracted representations, the need for labeled data is significantly reduced in model training.
    \item We propose a new loss function SME to solve the imbalanced dataset problem, which is shown to be efficient in experiments. SME can also be used by other unsupervised models that have imbalanced datasets.
    \item 
     We implement the proposed framework using reshape layers, assess its performance in diverse settings, and compare it to state-of-the-art learning-based approaches. We also compare the performance with other commonly used loss functions, i.e., MAE and MSE. 
    The results demonstrate the effectiveness of PowerBERT in learning effective representations for identifying FDIA and TDA. The results also indicate that the impact of TDA can propagate to a wider range, whereas the impact of FDIA is primarily limited to the area where the attack is launched and its directly connected areas.
    The code of implementation is now open-sourced\footnote {https://github.com/fridge23/PowerBERT}.
    % Our model was developed and experimentally evaluated. The results show that our model outperforms the random forest model in TDA classification by 5 \%, The classification results of the model trained with 0.05\% of the data outperform other deep learning models by 45\% in TDA classification and 25\% in FDIA classification, which demonstrated the effectiveness of our model in detecting and classifying time-delayed attacks and FDI attacks in the power grid. The code of our model is publicly available.
\end{itemize}
The rest of this paper is organized as follows. Section \ref{BG} introduces the system model and attack models. Section \ref{RW} discusses the related work. Section \ref{MT} presents the methodology and design of the framework. Section \ref{EE} reports the experiment settings, ablation study results and comparative performance of PowerBERT-based method and state-of-the-art methods. Section \ref{CC} concludes this paper.
\section{Related Work}
\label{RW}

% \subsection{Attack Detection}~\label{sec:systemmodel}
%Attacks in power systems significantly affect the working efficiency of the system, even making the entire system crash, and the delay attack and the FDI attack are two typical types.
Researchers have proposed signal processing based and machine learning based approaches to detect cyber attacks in smart grids.

\textbf{Signal processing based.}
% Some works \cite{tummalatwostageKF,dehghaniWSE} proposed to detect cyber attacks using the classic signal processing models, where the approaches such as Kalman filter and wavelet singular entropy are used for detecting the existence of cyber attacks. 
% \cite{tummalatwostageKF} presents a two-stage Kalman filter to detect cyber attacks and estimate the bias of the attacks.
% \cite{dehghaniWSE} uses the wavelet singular entropy in FDIA detection.
% These methods only perform anomaly detection without the ability to identify different types of attacks.
Some research works \cite{tummalatwostageKF,dehghaniWSE} have proposed to detect cyber attacks using classic signal processing models. These approaches use methods such as Kalman filters and wavelet singular entropy to detect the existence of cyber attacks.
\cite{tummalatwostageKF} presents a two-stage Kalman filter to detect cyber attacks and estimate the bias of the attacks.
\cite{dehghaniWSE} uses the wavelet singular entropy in FDIA detection.
However, these methods only perform anomaly detection and do not have the ability to identify different types of attacks.
% Several studies \cite{tummalatwostageKF,dehghaniWSE} have proposed using classic signal processing models to detect cyber attacks. For example, Tummala et al. \cite{tummalatwostageKF} present a two-stage Kalman filter to detect cyber attacks and estimate their bias, while Dehghani et al. \cite{dehghaniWSE} use wavelet singular entropy for FDIA detection. These methods can identify anomalies, but they cannot distinguish between different types of attacks.

\textbf{Machine learning based.}
% Compared with the signal processing based approaches, machine learning based approaches are more robust to the changes and noises in the environment.
% They include supervised learning approaches and unsupervised learning ones. Supervised learning based models \cite{Ganesh9,louxin12,detFDI25,FDILSTM26} have been proposed to detect various types of cyber attacks in smart grids. Lou et al.~\cite{louxin12} exploit BiLSTM based model for the detection of TDA. Mohammad Ashrafuzzaman et al. propose DNN based model~\cite{detFDI25} for FDIA detection. Qingyu Deng et al. use LSTM based model~\cite{FDILSTM26} to detect FDIA in a power grid. These approaches require a large amount of labeled data to train the model for accurate detection. 
% To reduce the reliance on labeled data, unsupervised learning approaches~\cite{anKNN,sakhniniUSL,uddinOCSVM,ahmedIF,HuSDAE,yangWPD} are studied. 
% \cite{sakhniniUSL} compares the performance of the combination of machine learning models and statistical feature extraction methods.
% \cite{anKNN} detects anomaly by leveraging KNN model.
% One-class SVM also be used in anomaly detection \cite{uddinOCSVM}, as well as isolation forest \cite{ahmedIF}.
% Yang at el. propose WPD-ResNet model to do transfer learning and detect anomalies in power station communication \cite{yangWPD}.
% And stacked denoising autoencoder is used to detect and classify several types of FDIA \cite{HuSDAE}.
% These methods either do anomaly detection \cite{anKNN,uddinOCSVM,ahmedIF} instead of attack classification or only target a specific type of attacks \cite{HuSDAE}.
Compared to signal processing-based approaches, machine learning-based approaches are more robust to changes and noise in the environment. They can be divided into supervised learning and unsupervised learning approaches.
Supervised learning-based models have been proposed to detect various types of cyber attacks in smart grids. For example, Lou et al.~\cite{louxin12} exploited a BiLSTM-based model for the detection of TDA. Mohammad Ashrafuzzaman et al. proposed a DNN-based model~\cite{detFDI25} for FDIA detection. Qingyu Deng et al. used an LSTM-based model~\cite{FDILSTM26} to detect FDIA in a power grid. However, these approaches require a large amount of labeled data to train the model for accurate detection.
To reduce the reliance on labeled data, unsupervised learning approaches~\cite{anKNN,sakhniniUSL,uddinOCSVM,ahmedIF,HuSDAE,yangWPD,dairi2023semi} have been studied. For example, \cite{sakhniniUSL} compares the performance of the combination of machine learning models and statistical feature extraction methods. \cite{anKNN} detects anomalies by leveraging a KNN model. One-class SVM has also been used in anomaly detection~\cite{uddinOCSVM}, as well as isolation forest~\cite{ahmedIF}. Yang et al. proposed the WPD-ResNet model to do transfer learning and detect anomalies in power station communication~\cite{yangWPD}. Stacked denoising autoencoders have been used to detect and classify several types of FDIA~\cite{HuSDAE}.
~\cite{dairi2023semi} proposed a semi-supervised learning model to do cyber-attack detection in smart grids by using a gated recurrent unit-based stacked autoencoder and a generative adversarial network model to learn the implicit features from the unlabeled data. It identifies anomaly data by using One-Class SVM as a binary classifier, so it can only do identification on one target attack, and the researcher trained different versions of models for various attacks.
However, these methods are either designed for anomaly detection~\cite{anKNN,uddinOCSVM,ahmedIF} instead of attack classification or only targeting a specific type of attacks~\cite{HuSDAE}.

In our earlier work~\cite{zengPowerBERT}, we proposed a BERT-like model to learn the generalizable and effective representations that capture distinctive patterns of different attacks 
% and, as a result, can be used to identify different types of attacks.
from the unlabeled data in real systems. However, this model was trained by using MAE as the loss function, which excelled primarily with balanced datasets – a challenging prospect to attain in actual systems. Furthermore, the model's evaluation was confined to a limited 3-area system, insufficient to fully explore the spatial nuances of attacks. To enhance the capacity of unsupervised learning models to extract meaningful features from imbalanced datasets, this paper introduces a novel loss function as an improvement upon our previous model. Additionally, we delve into a comprehensive spatial analysis of distinct attacks and assess detection performance within power grids spanning multiple areas, employing a more practical and extensive 5-area system.

\section{System model and attack models}
\label{BG}
In this paper, we consider the cyber attacks in automatic AGC in power grids as our case study for the proposed detection framework. In the following, we first introduce the AGC model and then describe attack models for FDIA and TDA, respectively.

% \subsection{Automatic Generation Control in Power Grids}\label{AGC}
% \begin{figure}
% \centering

% \includegraphics[width=0.4\textwidth]{bus.png}
% \caption{Three-area power grid system with 37 buses.}
% \label{bus}
% \end{figure}
\subsection{AGC Model}
\begin{figure}
	\centerline{ \includegraphics[width=0.48\textwidth]{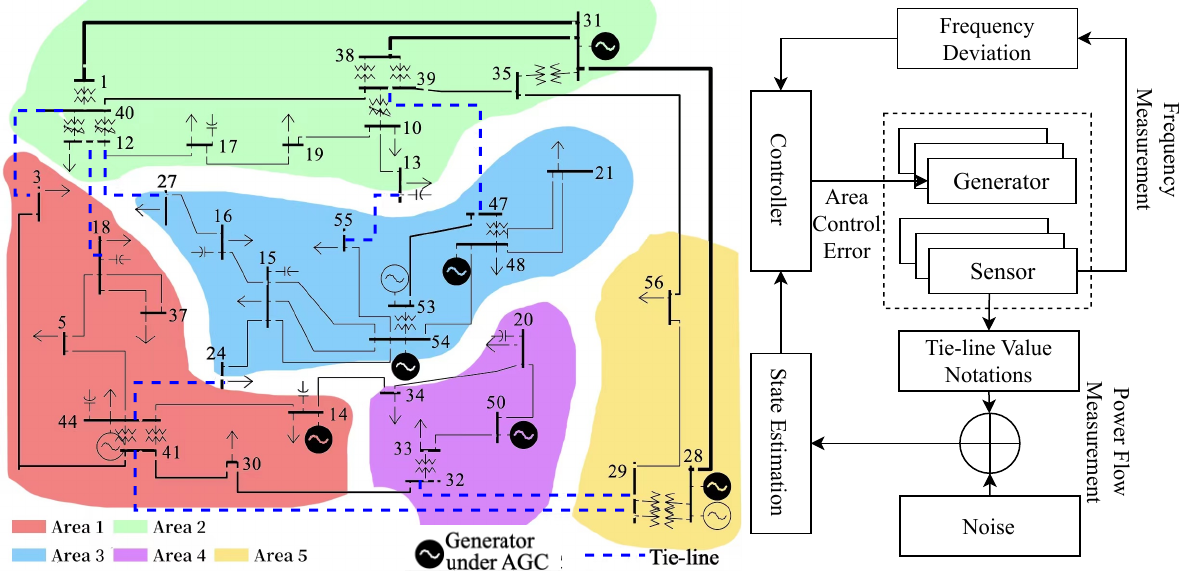}}
	\vspace{-0.1cm}
			\centerline{{(a)}\hspace{3.5cm} {(b)}}
	\caption{The system model. (a) Five-area power grid with 37 buses. (b) Overview of AGC.}\label{bus}
	\vspace{-0.3cm}
\end{figure}

% In a power system, AGC regulates the grid's frequency within a safe range by dynamically adjusting the system conditions in real-time~\cite{kundur14}. A power grid can be divided into several separate areas, and the AGC can also  control the power interchange rate among different control areas. In this paper, we discuss the discrete-time AGC system, where the time is divided into slots. We illustrate a three-area power grid with 37 buses in Figure~\ref{bus}(a)~\cite{Tan16}. This system involves five control areas and the dotted lines between two control areas are called tie-lines. In this paper, we use this 37-bus system as our case study to explore the cyber attacks in AGC control, which is a representative power grid model denoting a small to middle-scale real-world grid. 
The AGC system in a power grid dynamically adjusts the system conditions in real time to regulate the grid's frequency within a safe range~\cite{kundur14}. The AGC system can be divided into several separate areas, each with its own AGC controller. The AGC controllers communicate with each other to ensure that the grid's frequency is maintained within a safe range.
In this paper, we discuss the discrete-time AGC system, where the time is divided into slots. We illustrate a five-area power grid with 37 buses in Figure~\ref{bus}(a)~\cite{Tan16}. This system involves five control areas, and the dotted lines between two control areas are referred to as tie lines. We use this 37-bus system as our case study to explore cyber attacks in AGC control. This system is a representative power grid model that denotes a small to middle-scale real-world grid.

% In the AGC system, the area control error (ACE) is a control command used to regulate the generator output in the feedback control. For an area $i$ in the grid shown in Figure ~\ref{bus}(b), the command ACE$_{i}$ is a weighted sum of two signals inside the power grid i.e., the frequency deviation ($\Delta \omega_i$) and power export deviation ($\Delta P_{Ei}$). Thus, it can be  expressed as: $ACE_{i}=a_{i}\Delta P_{Ei}+b_{i}\Delta \omega_{i}$,
% where the $a_i$ and $b_i$ are two constant weights. The control center sends the ACEs command  to adjust the generator output via the communication network in different control area $i$. This control process is called the AGC cycle which is usually  about 2 to 4 seconds~\cite{kundur14}.
In the AGC system, the area control error (ACE) is a control signal used to regulate the generator output in a feedback loop. For an area $i$ in the power grid shown in Figure~\ref{bus}(b), the ACE$_{i}$ signal is a weighted sum of two signals within the grid: the frequency deviation ($\Delta \omega_i$) and the power export deviation ($\Delta P_{Ei}$). It can be expressed as follows: $ACE_{i}=a_{i}\Delta P_{Ei}+b_{i}\Delta \omega_{i}$, where $a_i$ and $b_i$ are two constant weights. The control center sends the ACE$_{i}$ signal to adjust the generator output via the communication network in control area $i$. This control process is known as the AGC cycle, and it typically takes 2 to 4 seconds~\cite{kundur14}.
% \begin{figure}
% \centering

% \includegraphics[width=0.35\textwidth]{bus.png}
% \caption{Three-area power grid system with 37 buses.}
% \label{bus}
% \end{figure}
% \begin{figure}
% \centering

% \includegraphics[width=0.4\textwidth]{AGC.pdf}
% \caption{Overview of AGC.}
% \label{AGC}
% \end{figure}
% \begin{figure}
% \centering

% \includegraphics[width=0.3\textwidth]{AGC.pdf}
% \caption{\label{agc}Overview of the AGC system structure.}
% \label{agc}
% \end{figure}
The power flow measurement from the power system is usually faulty and noisy, so the state estimation (SE) is designed to recover the information from a noisy signal. The measurement vector $y$ can be expressed as: $y=\textbf{M}x+\textbf{n}$, where  $\textbf{M}$ represents the measurement matrix, vector $x$ denotes all the states in the grid, and the $\textbf{n}$ denotes the noise. The target of SE is to estimate the state vector $x$ by $\hat{x} = (\textbf{M}^{\top}\textbf{W}\textbf{M})^{-1}\textbf{W}y$, where $\textbf{W}$ is a weighted matrix. Then the estimated power flow measurement is $\hat{y}=\textbf{M}\hat{x}$. In Bad Data Detection (BDD) ~\cite{liuBDD}, the alarm will be triggered if the difference between $y$ and  $\hat{y}$, i.e., $||y-\hat{y}||$, is bigger than a defined threshold.

% Fig~\ref{} summarizes the AGC system. {\red ******(description should be based on your diagram)}

\subsection{Attack Models}
In this paper, we use the representations learned using PowerBERT to detect two typical cyber attacks against AGC control in the power grid~\cite{lou20,Ganesh9}, the latest FDIA~\cite{yanSFDIA} and TDA~\cite{sargolzaeiTDA}. The learned representations can also be used to detect other types of cyber attacks in smart grids as long as they cause signal fluctuation in the system.

For traditional FDIA, after the adversaries know the power flow matrix $\textbf{M}$, they can add attack vector $a=\textbf{M}c$ into the power flow sensor measurement, where $c$ is an arbitrary vector, and the measurement becomes $\hat{y}=\textbf{M}(x+c)+\textbf{n}$, so BDD is bypassed because the noise does not change. The targeted FDIA~\cite{yanSFDIA} in our work not only lends matrix M to bypass BDD but also limits the magnitude of the false data added by the FDIA in a reasonable range so that the attack minimizes disruptions when it initially enters the system and keeps the frequency excursion long enough to ensure system damage. Compared to traditional FDIA, the attack we exploit is  stealthier and more destructive \cite{yanSFDIA}.

%For the time delay attack, the adversary's purpose is to delay the command of control from the controller by compromising the . Unlike the FDI attack, the adversary does not change the content in the communication packet, but just maliciously delays the packet by $\tau$ time slots. Thus, the generator receives the $t$th control command only at the $(t+\tau)$th time slot, which makes areas delay for $\tau$ slots and create the system frequency excursion. And sometimes the delay may also exist when there do not have any cyber attacks existed, because of the natural communication latency. In our paper, we assume that the clock of the controller and the actuator are asynchronous. Because the existing  clock  synchronization  protocols(e.g. NTP) also make the system become more vulnerable to attacks~\cite{louxin12,Ganesh9}. So we used a defense-in-depth paradigm to detect time delay attack and abandon all other orthogonal efforts to synchronize clocks securely in the CPS~\cite{Ganesh9}.

In the time delay attack (TDA), the adversary aims to delay the control command from the controller. Let $y(t)$ denote  the control command generated and transmitted by the control center in the $t$th time slot. The adversary maliciously delays these packets  by  $\tau$ time slots. Thus, in the $(t +\tau)$th time slot, the command $y(t)$ arrives at the actuator. Since we consider the  discrete-time  AGC control system in this paper,   the delay length $\tau$ is an integer. Moreover, different from FDIA, the adversary does not modify any content of the transmitted packet. The TDA can be launched by compromising  the  data communication  channels (e.g.,  compromised routers)  between  the  controller  and  the  actuator  to delay  the  transmission  of  control  commands~\cite{lou20}. Note that delayed signals may exist in the system even without the cyber attacks  due to the natural communication latency.  In the AGC, the attacker delays the control command in one of the areas $i$, i.e., ACE$_i(t)$,   by $\tau$ slots,  to create the system frequency excursion.  
% In this paper, we do not assume  the clocks of the controller and the actuator are synchronized. This is because the existing  clock  synchronization  protocols  are vulnerable to cyber attacks too~\cite{lou19, prakhar21}. Thus, a defense-in-depth scheme for the delay attack detection, which is independent of any clock synchronization in the power grid,  is indispensable~\cite{prakhar21}.

Overall, by either compromising the sensor readings (i.e., FDIA) or delaying the control commands (i.e., TDA), the purpose of the adversary is to make the system's frequency exceed the safety threshold and then force the disconnection between the generator and load or damage equipment. Same as the existing work~\cite{yanSFDIA,Ganesh9}, we consider the safety range of the frequency deviation as [-0.5,\;\;0.5] Hz, and the deviations out of this range are regarded as unsafe.

\section{Methodology}
\label{MT}

\begin{figure}
\centering

\includegraphics[width=0.5\textwidth]{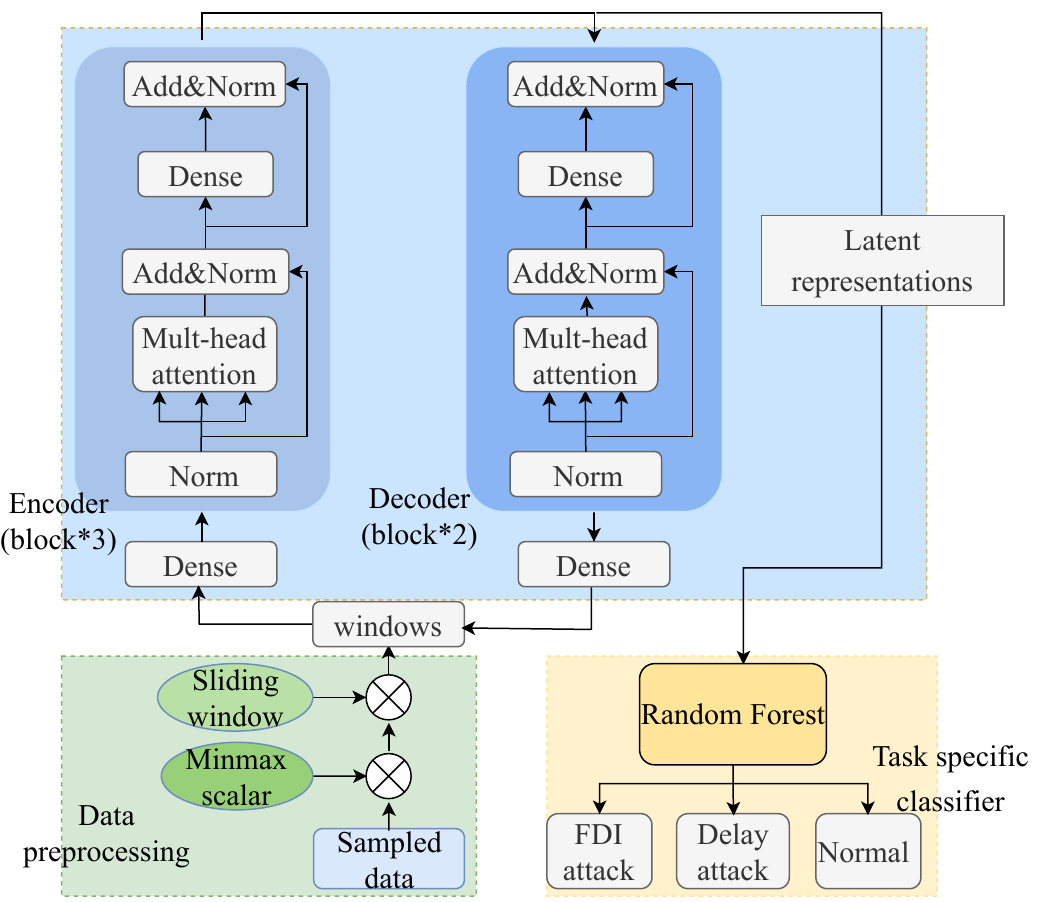}
\caption{\label{model}The proposed framework is comprised of data preprocessing, the PowerBERT model and a task-specific classifier.}
\label{struc}
\end{figure}

In this section, we introduce the details of the proposed cyber attacks detection and classification model. The overview of our framework is illustrated in Figure \ref{model}, which consists of 3 phases, i.e., data preprocessing, PowerBERT, and downstream classifier training. The sensing data collected from neighboring control areas in AGC are firstly normalized and extracted with a specific data structure. All the extracted sets of data are then fed into the PowerBERT self-supervised learning model to learn representations, which are used to train the downstream task-specific classifier with supervised learning.The details of this model is introduced below.
% \begin{figure}
% \centering
% \includegraphics[width=0.3\textwidth]{main structure.jpg}
% \caption{\label{fig:1}The framework of the whole model.}
% \label{framework}
% \end{figure}
\subsection{Data Preprocessing}\label{DP}

% The time-series data are sampled every four seconds in the simulation experiment, and each trace has 75 sample points and lasts for 5 minutes.
% Since 5 minutes is too long to protect the system from huge damage, we use sliding windows of size $WS$ to get data last for shorter time. And the label of the window is the attack's name if that attack sample is inside the window, or negative if there is no attack happened inside.
\textbf{Data normalization}: We use the min-max scalar to normalize the collected ACE data. The normalized data sample can be express as:$x_i'=\frac{x_i-x_{min}}{x_{max}-x_{min}}$,where the $x_i'\in(x_1',x_2'...,x_n')$ is the scaled result, $x_i\in(x_1,x_2...,x_n)$ is the original value and $n$ is related to the amount of data we have, $x_{min}$ is the smallest value and the $x_{max}$ is the biggest.
By using scalar, all the data are in the range of [0,\;\;1]. 

\begin{figure}
\centering
\includegraphics[width=0.45\textwidth]{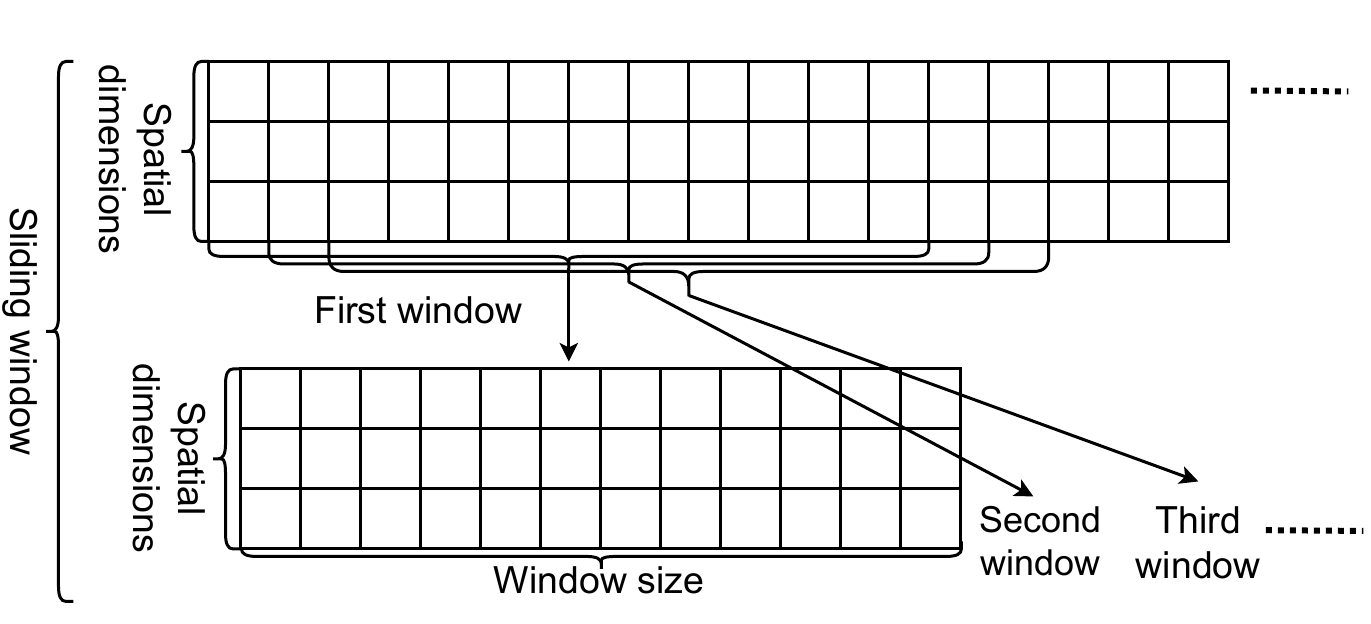}
\caption{\label{fig:event}The process of data extraction.}
\label{event}
\end{figure}

\textbf{Data extraction}: We use a sliding window to extract data clips from the normalized dataset. As illustrated in Figure \ref{event}, the sliding window with width $w_1$ is used to extract a set of data segments $B_i\in(B_1,B_2,...,B_m)$ from a data trace collected in the 5 neighboring areas, and $w_1$ is also the window size of attack detection.
% For each $B_i$ $(i=1,2,...,m)$, we divide the data segment with a second window $w_2$ into a set of data partitions $E_i={E_{i1},E_{i2}...E_{ik}}$, where each partition $E_{ij}$ $(j=1,2,...,k)$ captures the sensing signal distribution at a short time interval, and we regard such a partition as an event.
% All the events in $E_i$ capture the signal fluctuation in each detection window. 
% As illustrated in Figure \ref{event}, before $E_i$ is fed into PowerBERT, we use a reshape layer to reshape the data from $w_1\times5_{dim}$ to $\frac{w_1}{w_2}\times(5w_2)_{dim}$.

% From our observation, we find the The impact of the attack occurs gradually and is spatially transmissive. Instead of feeding the data individually, we combine some spatially and temporally adjacent data and input them together, and we regard the process as the event mechanism. To achieve this design, we reshape the data before feeding the data into the encoder. For instance, in the 3-areas AGC system, the original data can be expressed as $WS\times 3$, where $WS$ means the size of the sliding window, and for each sample, we have $3$ ACEs from $3$ different areas. We combine $EL$ sample points as an event, so we get the data of $\frac{WS}{EL}\times3EL$ size, and input this data into the encoder.

% The process of sliding window and event mechanism is shown in fig.~\ref{event}.

% Self-supervised Representation Learning
\subsection{PowerBERT}
% After data processing, we input the processed data into the model to do feature extraction and classification, the model is shown in Fig.\ref{model}.
% The model involved three main parts, namely the encoder, the decoder, and the classifier. For the encoder, we feed the processed unlabeled data into the encoder and output some high-level representation. Then the decoder reconstructs the original data based on the extracted feature. Once the model recovers the data favorably, we input the extracted features into the classifier to do classification.

PowerBERT is adopted from BERT model \cite{Devlin4,MAE,xu2021limu} to extract the high-dimensional representations from the massive unlabeled data. In NLP domain, BERT is a bidirectional model used to pre-train deep bidirectional representations from unlabeled text by jointly conditioning on both left and right context.
In this paper, we do not make use of the span mask algorithm to train PowerBERT. It is because the inputs in original BERT are word tokens while the input data in PowerBERT are continuous data samples.
The whole model is trained by back propagation to reduce the reconstruction error.

% Instead of using the masked design for our task, we considered that in the NLP, the data is discrete, every figure representing an individual word, while the data in our task is continuous, so the masked autoencoder may not be the best option. According to the experiment result, an autoencoder is chosen.

Before the processed data are fed into the encoder, a dense layer is used to embed data into high-dimensional tensor. For instance, the $ws$ points' data is embedded from $ws\times5_{dim}$ to $ws\times D_{dim}(D\geq5)$.

\textbf{Encoder}: The encoder involves 3 transformer blocks. Each block includes a layer Normalization layer, a multi-headed attention layer, an adding layer that works as a residual connection, and finally a fully connected layer. The process of block $i$ can be expressed as:
 $B^i=MultiAttn (LayerNorm (A_{in}^i)),A_{out}^i=Dense(LayerNorm(B^i+A_{in}^i))+B^i+A_{in}^i$,
% \begin{gather*}%不会产生编号
% B^i=MultiAttn (LayerNorm (A_{in}^i)),\\
% A_{out}^i=Dense(LayerNorm(B^i+A_{in}^i))+B^i+A_{in}^i,\nonumber
% \end{gather*}
where $A_{in}^i$ denotes the data fed into block $i$, $B^i$ denotes the data output by multi-headed attention layer, $A_{out}^i$ denotes the output data of block $i$.

\textbf{Decoder}: The outputs of the encoder go to the decoder, where the extracted high-dimensional representations are reconstructed. The decoder has 2 blocks inside. The encoder has more blocks than the decoder, for the reason that we need a more complex encoder to extract better features for the downstream tasks. The formulas of the blocks $m$ can be expressed as: $D^m=MultiAttn (LayerNorm (C_{in}^m)),
C_{out}^m=Dense(LayerNorm(D^m+C_{in}^m))+D^m+C_{in}^m$,
% \begin{gather*}%不会产生编号
% D^m=MultiAttn (LayerNorm (C_{in}^m)),\\
% C_{out}^m=Dense(LayerNorm(D^m+C_{in}^m))+D^m+C_{in}^m\nonumber,
% \end{gather*}
where $C_{in}^m$ is the inputs of block $m$, and $C_{out}^m$ is the outputs of block $m$, and $D^m$ denotes the outputs of multi-headed attention layer.
After the decoder, a fully connected layer is designed to reshape the data back to the original structure.

\textbf{Train}:
% The loss of the autoencoder is computed based on the difference between the original data and the reconstructed data. By using backpropagation, the weights in the model are updated, whereas Adam optimizer \cite{kingma2014adam} is used for updating weights. For the learning rate, we use the learning rate warm-up to speed up the training process.
The autoencoder loss is computed as the difference between the original data and the reconstructed data. The model weights are updated using backpropagation, while the Adam optimizer \cite{kingma2014adam} is used to update the weights. The learning rate is warmed up to accelerate the training process.

\textbf{Loss function}:
% For the reason that the data is not balance in
To address the issue of imbalanced datasets in unsupervised learning for real-world applications, where the DL model can easily overlook large errors in the category with a small amount of data, we propose a new loss function called Separate Mean Error (SME). The SME loss function aims to give equal consideration to all categories. The expression of the loss function is shown below,
\begin{gather*}%不会产生编号
SME=Mean(s_1,s_2,s_3,...,s_n)+Mean(l_1,l_2,l_3,...,l_m),
\end{gather*}
where we divide all the training errors into two groups, the larger error group $(l_1,l_2,l_3,...,l_m)$ and the smaller error group $(s_1,s_2,s_3,...,s_n)$, by using the adjustable threshold $k$, and the SME is the sum of the mean values of those two groups.

If the degree of data imbalance is unknown, we recommend using SME with the average value of the entire batch as the threshold for training. Because the model will focus on dragging the large errors and small errors in the batch closer to a value (i.e., the mean of large and small errors) while training. It can help the model learn more comprehensive features and information from the imbalanced dataset.

\textbf{Feature extraction}: After training, we extract the appropriate features from PowerBERT for the training of the downstream classifier as illustrated in Figure \ref{struc}.
In contrast to our previous work~\cite{zengPowerBERT}, the new model exhibits a streamlined feature extraction approach. We have foregone the utilization of the Gaussian Mixture Model for uncovering latent insights within the auto-encoder's reconstruction error. This adjustment stems from the adoption of the SME function, which inherently captures this pertinent information during the model's advanced training process.

%In addition to the latent representations, we also make use of the reconstruction error to better train the classifier. The representations are the outputs of the encoder of PowerBERT, and the reconstruction errors, , are the reconstruction errors of PowerBERT, i.e., $X_{in}-X_{out}$, where $X_{in}$ is the input of PowerBERT, and $X_{out}$ is the reconstructed results. 
%{\color{red}{The reconstruction errors of the labeled data are directly used as features in the downstream classification task.}}

\subsection{Downstream Classifier Training}

% Once the features have been well learned from PowerBERT, we use a small amount of labeled data to train the classifier to do the classification of FDIA and TDA. 
% We deploy a random forest model with 1000 estimators as the classifier. Although being a small amount, the labeled data includes all the targeted types of cyber attacks. 
% In our case study, the classifier is trained to identify the data without an attack (i.e., normal), FDIA attack and TDA.
After the features have been well learned from PowerBERT, we use a small amount of labeled data to train a classifier to distinguish between FDIA and TDA. We deploy a random forest model with 1,000 estimators as the classifier. Although being a small amount, the labeled data includes all the targeted types of cyber attacks.  In our case study, the classifier is trained to identify data without an attack (i.e., normal), FDIA attacks, and TDA attacks.

\section{Evaluation}
\label{EE}
We now evaluate the performance of the proposed framework for detecting the FDIA and TDA against AGC in the power grid. We first
describe the dataset and evaluation metrics, and then briefly introduce other state-of-the-art attack detection models. After that, we show our model performance and the comparison with other models.

\subsection{Methodology}
\textbf{Dataset}: We use the industry-strength power system simulator PowerWorld \cite{pw} to simulate cyber attacks against AGC in a five-area 37-bus model as shown in Figure \ref{bus}(a).  We add randomly generated load deviations to simulate real-world dynamics.  
The ACE data samples are collected every $4$ seconds. All the attacks are launched in one of the control areas. In this paper, we choose to launch the TDA (the length of the delay is between 1 and 20 slots) in the generators on bus 14 in Area 1 and launch  FDIA in the tie-line from Bus 29 to Bus 41, and we can get similar results if the attack is launched in any other control areas too. We collect data from the five control areas shown in Figure \ref{bus}(a) when the power system is under FDIA \cite{yanSFDIA}, TDA \cite{sargolzaeiTDA}, and without attack, respectively, and all the attacks are launched in one of the areas at a random time. If the extracted data segment (as introduced in Section \ref{DP}) contains any data samples that are collected when the system is under attack, the segment is labeled as the corresponding type of attack. 
In total, we collect around 11861 traces, with 5285 traces without any attack, 3416 traces involving TDA, and 3160 traces involving FDIA. 

We divid the data into training $(43\%)$, validation $(7\%)$, and testing $(50\%)$ set. To simulate the real-world scenarios, the validation set and training set only involve  a few attack data but a large amount of normal data. They involve normal data $(99.98\%)$, and two types of attack data $(0.01\%)$. In the testing set, the different categories' data are balanced.

\textbf{Metrics}: We use the precision, recall and F1-score to evaluate the model performance. Specifically, $
Precision= \frac{TP}{TP+FP} $, $
Recall=\frac{TP}{TP+FN} $, $
f_1-score=\frac{Precision+Recall}{2} 
$, where $TP$ denotes true positive, meaning the data segment is classified as the correct class; $TN$ denotes true negative, meaning the data segment of other classes is not classified into the class; $FP$ denotes false positive, meaning the data segment of other classes is classified as the class, and $FN$ denotes false negative, meaning the data segment is classified as other classes.

% \begin{table}
% \centering
% \begin{tabular}{|l|r|r|r|}
% \hline
% &normal trace&FDI trace& delay trace \\\hline
% quantity & 6944&4990&5000 \\
% \hline
% \end{tabular}
% \caption{\label{struc}The data structure inside the dataset.}
% \end{table}

% Next, randomly choose $70\%$ of traces out of the whole data-set as the train data-set, and the rest are the test data-set. The train data-set is used for auto-encoder training, and randomly choose $8000(1\%)$,$800(0.1\%)$,$400(0.05\%)$ windows in the train set with their labels as classifier trainset, each classifier training set involves $50\%$ of normal data, $25\%$ of time delay attack data, and $25\%$ of FDI attack data, and the classifier is tested by the test data-set. For classifier training and auto-encoder training, we select $15\%$ of data in the relevant train set for validation.
\subsection{Different Learning Models}\label{MC}
In the evaluation, we compare the proposed model with other alternative models, which are based on state-of-the-art machine learning models in the literature.

~\textbf{DNN model}~\cite{detFDI25}: It is a MLP model, which involves 3 hidden layers. Because it was only designed for FDIA detection, so we change the last layer of the model from 2 units to 3 units and train it to do classification task.

~\textbf{RNN model}~\cite{FDILSTM26}: RNN model is very sensitive with the temporal information. We used an RNN model with 3 LSTM layers which have 64 units and a 33-unit fully connection layer. For the output layer, we set 3 units to classify the data into different categories.

\textbf{DB-RF}~\cite{farrukhDBRF}: A variant of random forest, which involves two random forest levels, and the first level performs anomaly detection, and the second level identifies the type of attacks. Two levels work with different kinds of features. We set the model with 330 estimators and train it to do triple classification.

\textbf{RF}: A random forest model that is trained with the raw data instead of the learned representations. The model has 1000 estimators to identify the types of attacks.

\textbf{PB+RF model~\cite{zengPowerBERT}}: The old version of PowerBERT we proposed in~\cite{zengPowerBERT} to extract representations, and a 1000 estimators' random forest classifier and identify data into 3 categories.

\textbf{PB+RF model}: We use the new PowerBERT proposed in this paper to extract representations, and a 1000 estimators' random forest classifier and identify data into 3 categories.

% To show the effectiveness of representations learned by our model, we used a random forest model with 1000 estimators to classify data into 3 categories.

PowerBERT and other models are implemented with python, scikit-learn and tensorflow~\cite{tensorflow,sklearn}. They are trained in a Google Colab server with T4 GPU. The learning rate and batch size in both self-supervised and supervised training phases are 1024.

\subsection{Evaluation Results}
% We set 3 different sliding window sizes, 30, 15, and 5. We compare the performance of different models with different window sizes to prove that our model is capable to do fast detection with a few labeled data. And because there have three different ACEs from different areas, our data has three  dimensions for each sample point, and the input data structure would be $30_{SP}\times3_{dim}$.
\subsubsection{Sliding window size}
\begin{figure}
\centering
\includegraphics[width=0.5\textwidth]{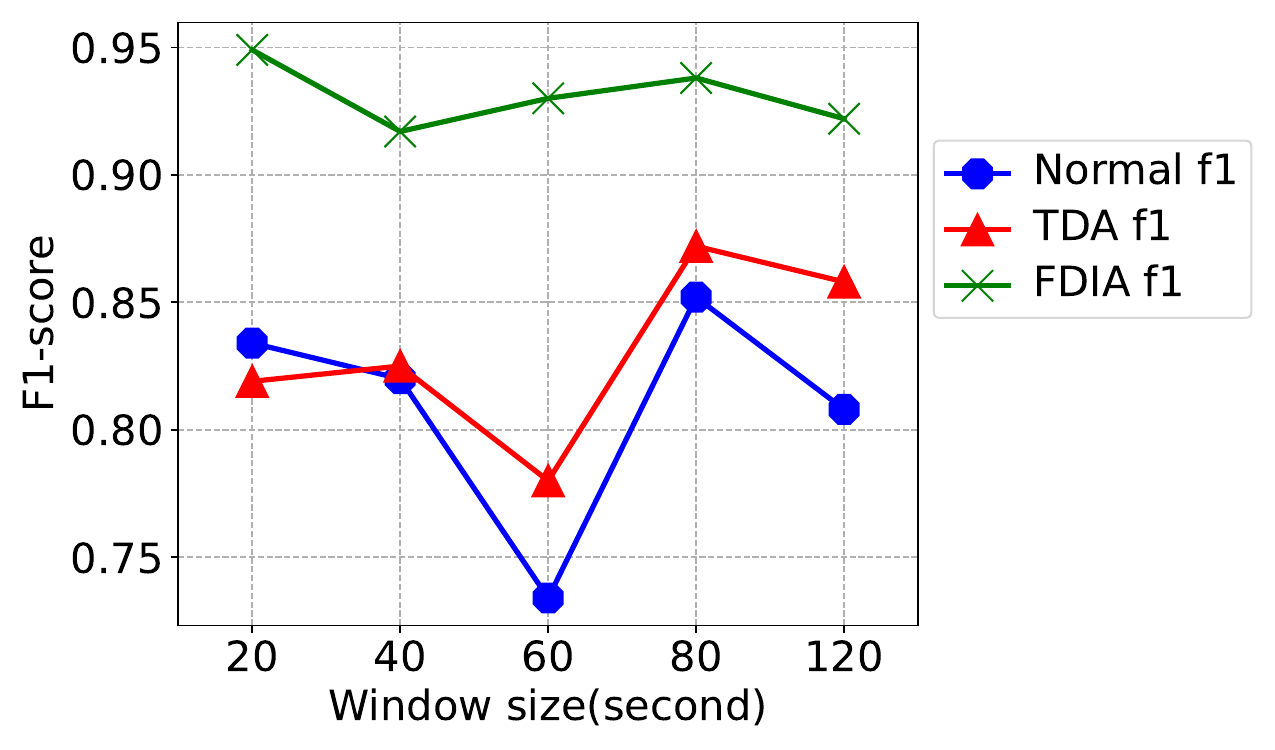}
\vspace{-0.3cm}
\caption{\label{WS comparison}Performance illustration of PowerBERT+RF under different sliding window sizes.}
\label{WS comparsion}
\end{figure}
% \begin{table}
% \caption{Experiment result comparison for event window size selection.}
% \centering
% \begin{tabular}{|c|c|c|c|}
% \hline
% % \multirow{2}{*}{1} &\multicolumn{3}{c|}{400 windows}\\\hline
% Event size(s)&Normal f1&TDA f1& FDIA f1 \\\hline
% 4&\textbf{85.0\%}&\textbf{86.4\%}&\textbf{94.1\%}\\\hline
% 8&83.1\%&83.7\%&93.7\%\\\hline
% 16&73.9\%&79.5\%&91.4\%\\\hline
% 20& 77.1\%&82.6\%&88.9\% \\\hline
% 40&66.4\%&77.0\%&87.8\%\\\hline
% \end{tabular}
% \vspace{-0.3cm}
% \label{event result}
% \end{table}

We test the performance of our model with different sizes for the sliding window $w_1$. Figure \ref{WS comparison} plots the F1-score of the models with $w_1$ size of $20, 40, 60 ,80$ and $120s$. We see that as the window size increases, the model performs best in TDA identification when the window size is $20s$, and performs best in FDIA at window size $80s$. In order to reduce the computation overhead and ensure prompt detection for different categories, we use $w_1=80s$ in the following performance evaluations.

% From the table we can find that the model has the best performance when the event size is 5, $4.5\%$ outperformed no event design. For the rest of the experiment, we use this design for comparison.
% \subsubsection{Event window size}
% As introduced in Section \ref{MT}, in each Sliding window, we divide the data samples into data partitions with window size $w_2$, and regard the partitions as \textit{events}. 
% We compare the performance of the models with different event window sizes $(w_2=4, 8, 16,20,40 s)$ for data partitioning in Table~\ref{event result}. 
% {\color{red}{Since window size $w_1=80s$, which corresponds to 20 data samples collected from each grid area, each event window $w_2$ involves  $(20,10,5,4,2)$ \textit{events}, respectively.
% As presented in Table \ref{event result}, the model with \textit{event} window size of $4s$ performs the best, which corresponds to 1 sample from each area. This is different from the result in the prior work~\cite{zengPowerBERT} for the 3-area 37-bus system, where a larger event size (i.e., $w_2=20$) acheives better performance. The reason is.

% %In another word, the event design is not suitable for the 5-area 37-bus smart grid system, although it worked for the 3-area 37-bus smart grid system~\cite{zengPowerBERT}.

% }}

\subsubsection{Threshold k in SME}
\begin{table}
\caption{Performance comparison of PowerBERT+RF with different SME threshold.}
\centering
\begin{tabular}{|c|c|c|c|c|}
\hline
% \multirow{2}{*}{1} &\multicolumn{3}{c|}{400 windows}\\\hline
Threshold $k$ &Normal F1&TDA F1& FDIA F1 \\\hline
$Mean(te)\times2$&83.2\%&87.2\%&89.1\%\\\hline
$Mean(te)\times1.5$&\textbf{85.9\%}&\textbf{87.2\%}&\textbf{93.8\%}\\\hline
$Mean(te)\times1.2$&79.7\%&82.2\%&93.3\%\\\hline
$Mean(te)$&84.1\%&86.5\%&92.6\%\\\hline
$Mean(te)\times0.8$&68.8\%&76.2\%&92.8\%\\\hline
$Mean(te)\times0.5$&84.5\%&85.3\%&92.9\%\\\hline
\end{tabular}
\vspace{-0.3cm}
\label{threshold}
\end{table}

As we mentioned in Section.~\ref{MT}, we propose a new loss function SME, where the training errors are divided into two groups based on a threshold $k$. We test our model trained with SME as the loss function under different thresholds $k$, where $k$ is set as  $2, 1.5, 1.2, 1, 0.8, 0.5$ times the mean value of all the training errors, i.e., $Mean(te)$. The model F1-score is shown in  Table.~\ref{threshold}. We can find that the model performs best when the threshold is 1.5 times the training error mean value $Mean(te)$. Thus, in all of the following experiments, we set the threshold $k$ in SME as 1.5 times $Mean(te)$.
However, for other datasets that are more balanced, SME with a threshold closer to $Mean(te)$ can get better performance.

\subsubsection{Impact of spatial effect}
\begin{table}
\caption{Performance comparison of PowerBERT+RF using the data from individual control area and all 5 control areas.}
\centering
\begin{tabular}{|c|c|c|c|c|}
\hline
% \multirow{2}{*}{1} &\multicolumn{3}{c|}{400 windows}\\\hline
Measurement availability &Normal F1&TDA F1& FDIA F1 \\\hline
Area 1&68.7\%&83\%&70.0\%\\\hline
Area 2&45.2\%&73.4\%&47.9\%\\\hline
Area 3&58.7\%&73.0\%&89.6\%\\\hline
Area 4&50.4\%&76.9\%&43.6\%\\\hline
Area 5&59.0\%&77.2\%&67.1\%\\\hline
Area 1,3,5&77.7\%&80.5\%&\textbf{95.3\%}\\\hline
All areas&\textbf{85.9\%}&\textbf{87.2\%}&93.8\%\\\hline
\end{tabular}
\vspace{-0.3cm}
\label{N&I}
\end{table}

We consider three different settings here: 1) The measurements collected from all five control areas; 2) The measurements collected from the area where attacks are launched and the directly connected areas; 3) The measurements connected from one of the five control areas.  Table \ref{N&I} presents the performance of PowerBERT+RF under these settings. Our results demonstrate that the F1-score is significantly higher when measurements from all five control areas are available. Specifically, the F1 scores for Normal, TDA, and FDIA all exceed 85\%, which illustrates the effectiveness of using spatial redundancy in smart grids. When we examine individual area measurements, we observe that our model exhibits better performance when the measurements are obtained from Area 1 under TDA, compared to other areas. The F1-scores are approximately 10\% higher in this scenario,  owing to the fact that the attack involves delaying the command that reaches the generator in Area 1, which misleads the generator output in Area 1. As a result, the TDA has the greatest impact in Area 1, which makes it easier for PowerBERT+RF to detect the TDA in Area 1. Moreover, since all areas are interconnected, the effects of the attack propagate to other parts of the grid as well, which enables PowerBERT+RF to achieve relatively good performance in other areas as well. With regards to FDIA, PowerBERT+RF achieves higher F1-scores when the measurements are from Areas 1, 3 and 5, respectively. Conversely, if the measurements are taken from Area 2 or Area 4, both F1-scores are below 50\%. The reason is that the AGC command generated at Area 1 using the compromised tie-line values directly affects the power balance between Area 1 and its connected areas, i.e., Areas 3 and 5.  These findings suggest that the impact of TDA can propagate to a wider range, whereas the effects of FDIA are primarily limited to the attack launch area and its directly connected areas.  

Based on this observation, we train the model by using the measurements from the attack launch area and its directly connected areas, to compare the performance with the case that measures are from all five areas. From the results, we can find that if the measurements are from all five areas,  PowerBERT+RF still achieves better performance when there is no attack or the attack is TDA.  However, the model's performance is worse when the attack is FDIA. This demonstrates that the impact of TDA spreads to a wider range, allowing the model to extract more information from all five areas compared to just three areas. In contrast, the impact of FDIA is limited to the three areas, and measurements taken from Areas 2 and 4 may not be helpful and can even negatively impact the identification process.

%While for the settings that only the individual area measurement is available, detection performance for all three attack scenarios is better if the measurement is only from Area 1, i.e.,  detection performance  for Nomal, TDA and FDIA are at least around 10\%, 6\% and  higher than that of other areas. The FDIA detection fo

%We compare the performance of two different versions of our model trained using the ACE data from individual control areas and all 5 control areas, respectively. The results are reported in Table \ref{N&I}, where we provide the performances (all attacks are launched in area1) of single-area PowerBERTs among the 5 control areas respectively and PowerBERT. We find that the performance of PowerBERT is significantly better than single-area PowerBERT, which demonstrates the effectiveness of using spatial redundancy in smart grids. And for the individual area PowerBERT, PowerBERT for area 1 gets the best performance for TD attack identification, less than 10\% higher than other areas, while the performance of FDI attack gets an extreme difference at more than 40\%, which means that the attack-launched area gets the largest impact from both the TD and FDI attack, but the spatial impact of TD attack is much larger than FDI attack. By observing the experiment results, it seems that the FDI attack only affects the attacked area and an area directly adjacent to it, while the TD attack has a significantly wider spatial impact.

\subsubsection{Effectiveness of SME}
We now compare the detection performance of PowerBERT trained by SME, MAE, and MSE, respectively. We use the same training set and validation set but the above three loss functions to train PowerBERT, and compare the performance with the labeling rate at 0.002\%. 
The comparison result is shown in Table~\ref{LF}. From the table, we can see that SME outperforms MAE and MSE in all three categories. SME's superiority lies in its focus on reducing the mean value of training errors while also paying attention to categories with larger training errors, which helps SME achieve better performance with imbalanced datasets during unsupervised learning.
On the other hand, models trained by MAE and MSE perform relatively well on two specific types of attacks but poorly on normal data. This suggests that these models have a weaker ability to differentiate normal data and attack data. The features learned by these models mainly pertain to normal data, which is insufficient to detect attacks. Although the F1 scores for specific attacks are relatively high, it is because the difference between TDA and FDIA is significant enough to distinguish them based on the information learned from a small amount of unlabeled data.
% Compared to the model trained by SME, the models trained by MAE and MSE perform relatively well on the two specific types of attacks but poor on normal data, which means those models have a weak ability to differentiate normal data and attack data but have the better ability on identifying two types of attacks. Because the features learned by those models are mainly about normal data which are insufficient to detect attacks, the F1-score for specific attacks are relatively high for the difference between TDA and FDIA is significant enough for us to be able to distinguish them based on the information learned from a small amount of unlabeled data.

\begin{table}
\caption{Performance comparison of PowerBERT+RF training with different loss functions.}
\centering
\begin{tabular}{|c|c|c|c|c|}
\hline
% \multirow{2}{*}{1} &\multicolumn{3}{c|}{400 windows}\\\hline
Loss Function &Normal F1&TDA F1& FDIA F1 \\\hline
SME&\textbf{85.9\%}&\textbf{87.2\%}&\textbf{93.8\%}\\\hline
MAE&70.7\%&77\%&91.1\%\\\hline
MSE&71.1\%&77.3\%&91.3\%\\\hline

\end{tabular}
\vspace{-0.3cm}
\label{LF}
\end{table}
% \subsubsection{Effectiveness of reconstruction error distributions}
% \begin{table}
% \caption{Performance comparison for classifiers with different features.}
% \centering
% \begin{tabular}{|c|c|c|c|c|}
% \hline
% \multicolumn{2}{{|c|}}{Metrics}&Representation&Representation&PowerBERT\\
% \multicolumn{2}{{|c|}}{}&+RF&+mean+RF&+RF\\\hline
% 	               \multirow{3}[0]{*}{\rotatebox{90}{F1-score}}      &  Normal  &  96.4\%&96.4\%&\textbf{96.5\%}\\\cline{2-5}
% 	 &TDA&77.8\%&77.8\%&\textbf{78.8\%}\\\cline{2-5}
% 	 &FDIA&97.9\%&97.9\%&\textbf{98.0\%}\\\hline
% \end{tabular}
% \label{error}
% \end{table}

% We compare the performance of three classifiers trained with different feature settings using the labeling rate of 0.005\%.
% The classifiers include \textbf{representation+RF}: the classifier using the representations as features and \textbf{PowerBERT+RF}: the classifier using the representations and reconstruction error as features.
% % The comparison result is shown in table\ref{error}.Compared with other classifier, the GMM+RF get the best performance but also the highest complexity.
% As shown in Table \ref{error}, adding the mean value of reconstruction errors does not result in any performance improvement. By combining the error distributions and the representations, we can increase the detection F1-score by 1\% for the TD attack. Since the TD attack is one of the most difficult types of attacks to detect in practice, we employ the combination of representations and reconstruction error distributions as the features in our final framework design.

\subsubsection{Model comparison}
\begin{table}
\caption{Model comparison using different amount of labeled data for training.}
\centering
\begin{tabular}{|c|c|c|c|c|c|c|c|}
\hline
Labeled data &F1- & DNN &RNN& DB-&RF&PB&PB \\ 
portion (\%)&score &&&RF&&+RF\cite{zengPowerBERT} &+RF\\\hline
	               \multirow{3}[0]{*}{0.002\%}     &  Normal  &  18.5&49.4&50.9&56.9&77.7&\textbf{85.9}\\\cline{2-8}
	 &TDA&44.6&21.6&72.1&72.8&82.9&\textbf{87.2}\\\cline{2-8}
	 &FDIA&54.2&22.9&77.1&80.1&89.6&\textbf{93.8}\\\hline
	 \multirow{3}[0]{*}{0.008\%} &  Normal &  41.5&14.9&77.9&84.4&88.0&\textbf{90.6}\\\cline{2-8}
	 &TDA&11.4&21.9&86.7&86.6&86.4&\textbf{89.9}\\\cline{2-8}
	 &FDIA&41.3&38.3&76.6&88.0&95.9&\textbf{95.9}\\\hline
	 \multirow{3}[0]{*}{0.02\%}&  Normal&  51.3&25.0&88.6&89.3&91,1&\textbf{92.4}\\\cline{2-8}
	 &TDA&11.3&45.0&89.7&88.6&89.5&\textbf{91.9}\\\cline{2-8}
	 &FDIA&56.1&51.4&92.6&94.6&\textbf{96.2}&\textbf{96.2}\\\hline
\end{tabular}
\label{comparison result}
\end{table}
% \subsubsection{Model Comparison}
% \begin{table}
% \centering
% \begin{tabular}{|l|r|r|r|r|r|r|r|r|r|}
% \hline
% 	 & \multicolumn{3}{|c|}{400} & \multicolumn{3}{c|}{800}& \multicolumn{3}{c|}{8000} \\ \cline{2-10}
% 	                model      &  nof1  &   def1&FDf1&  nof1  &   def1&FDf1&  nof1  &   def1&FDf1 \\  \hline
% 	DNN& 93.67\%&58.67\%&85.67\%&95.33\%&67.67\%&91.67\%&96\%&77\%&97.67\%\\\hline
% 	RNN&94.67\%&57.67\%&94\%&96\%&72.33\%&96.67\%&96.5\%&80.5\%&98.5\%\\\hline
% 	Double RF&96\%&70.33\%&96\%&96.33\%&74\%&97\%97\%&81\%&98.33\%\\\hline
% 	RF&96\%&72.67\%&96.33\%&96.33\%&75\%&98\%&97\%&81.67\%&99\%\\\hline
% 	PowerBERT&96.2\%&76.04\%&97.63\%&96.45\%&78.38\%&97.99\%&96.89\%&83.01\%&98.58\%\\\hline
% \end{tabular}
% \caption{The test result for comparing models trained by different size of data.}
% \label{comparison result}
% \end{table}

In this subsection, we compare the detection performance of our model with other state-of-the-art models as introduced in Section \ref{MC} and our earlier work in~\cite{zengPowerBERT}. We show the performance of all the models with three different settings, where the amount of labeled dataset for model training is different. We use the labeling rate of 0.002\%, 0.008\%, and 0.02\% to train all the models respectively and compare their performance.
% We mainly compare the performance of models trained by 8000,800, and 400 labeled windows. The results is shown as Chart.~\ref{comparison result}. 

Table \ref{comparison result} summarizes the F1-score of attack detection using different models. PowerBERT+RF achieves the best performance in almost all settings, especially when the labeling rate is low. State-of-the-art DL and ML models suffer from very low accuracy due to the lack of enough labeled data for training.

%, and the increase of the F1-score for the attack data is more significant than the normal data because of the dramatic increase of the recall value
When comparing our earlier work~\cite{zengPowerBERT}, we see from Table~\ref{comparison result} that the previous model cannot perform as well as the new one in handling imbalanced datasets, particularly with low labeling rates. 
It is noteworthy that at a  labeling rate of 0.02\%, the model in earlier work~\cite{zengPowerBERT} performs similarly in FDIA identification since it doubles the computation overload to extract some information from the reconstruction error. However, we abandoned this design in our current work as its improvement is limited, and the new PowerBERT can still achieve a slightly better performance even with this design.
% Compared with other competitors, our model achieve the best performance in all aspects When the amount of label data is 400 and 800. Specifically, our model get at least 1.3$\sim$3.4\% improvement comparing other existing model in classification. When the labeled data size comes to 8000, FDIA classification result is 0.4\% lower than random forest, but still 1.3\% higher in TDA classification.
% \ref{}
% \begin{figure}
% \centering

% \includegraphics[width=0.5\textwidth,height=0.3\textwidth]{comparison.jpg}
% \caption{\label{comparison}The comparing result for 4 model with different window size and data size in classification of attacks.}
% \label{comparsion}
% \end{figure}
% From the folding Line Chart shown in Fig.~\ref{comparison}, 

% Our model has the best performance for almost all the conditions. The PowerBERT model gets better performance with the enlarging window size and data size. The PowerBERT model can get an f1 score of 80\% in TDA classification and 98.6\% in FDIA classification at 400 windows$(0.05\%)$ and window size 30. For shorter window size, it still gets a 70\% plus f1 score in TDA and 97\% plus in FDIA with the same data size. And our model is 3\~5\% better than the baseline model in TDA classification and 1\% better in FDIA, with 50\% improvement in TDA and 25\% improvement in FDI compared to deep learning models, which proves that our model can learn useful features from massive data and do the better downstream classification of cyber attacks with a limited size of labeled dataset.
\subsubsection{Representation visualization}

\begin{figure}
\vspace{-0.4cm}
\centering
\includegraphics[width=0.5\textwidth]{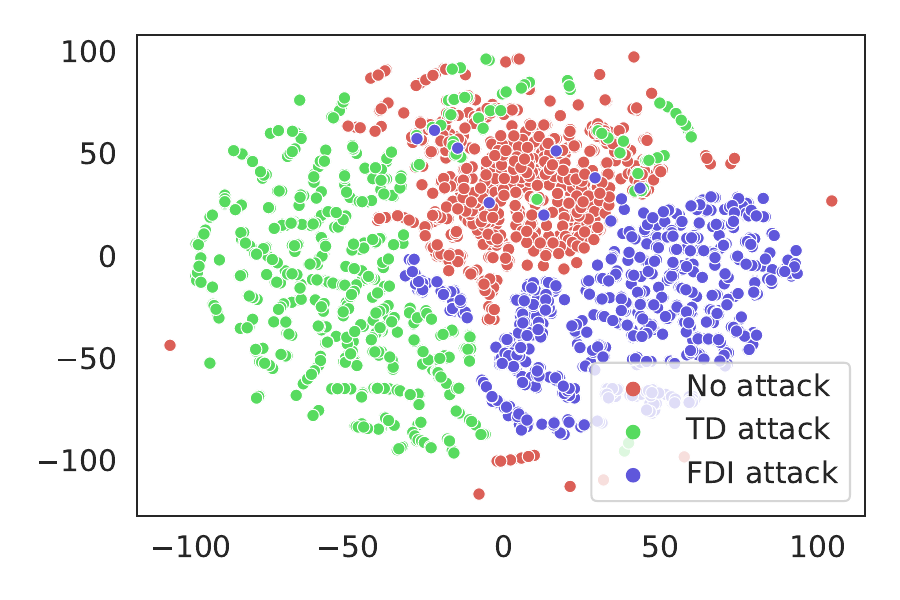}
\vspace{-0.3cm}
\caption{\label{comparison}Representation visualization with t-SNE.}
\vspace{-0.3cm}
\label{visual}
\end{figure}
To gain a more intuitive understanding of the effectiveness of the representation learned by our model in the classification task, we visualize the learned high-dimensional representations of data in 2D space by t-distributed stochastic neighbor embedding (t-SNE)~\cite{van2008visualizing}. 
We randomly select a total of 3000 equal amounts of no attack data, TDA data and FDIA data. Then they are feature extracted by PowerBERT, reduced to two-dimensional data with t-SNE and drawn on a scatter plot. The result is shown in Figure \ref{visual}.
% , where \textit{comp-1} and \textit{comp-2} are the two coordinates of the t-SNE mapping to the two-dimensional space.
It is obvious that samples belonging to the same types of attacks exhibit a high clustering effect after the representation extraction of PowerBERT.

\subsubsection{Computation overhead}
We show the inference speed of our model and other models in this part. For each model, we let it perform the attack detection on the test dataset and calculate the average detection time needed for each data clip, starting with feature extraction until we get the classification results from the ML model.
All the experiments are done on the Google Colab with T4 GPU. The results are reported in Table \ref{complexity}. Our findings demonstrate that our model can achieve real-time detection. Moreover, when comparing the inference time with the earlier work~\cite{zengPowerBERT}, we observe  that the detection time of the new PowerBERT model is halved, making it even more feasible for workstations to achieve real-time detection.
% We can find because our model do feature extraction and classification, the inference time is longer than other models, and DNN model have the shortest inference time.
\begin{table}
\caption{The inference time for each sample in different models (s).}
\centering
\begin{tabular}{cccccc}
\hline
DNN&RNN&DB-RF&RF&PB+RF~\cite{zengPowerBERT}&PB+RF\\\hline
8.20E-5&1.22E-4&7.54E-5&4.11E-5&4.17E-4&2.10E-4 \\\hline
\end{tabular}
\vspace{-0.3cm}
\label{complexity}
\end{table}
\section{Conclusion}~\label{CC}
In this paper, we proposed PowerBERT, a self-supervised learning model to learn the generalizable features from massive unlabeled sensing data for cyber attack detection in smart grids. We demonstrated the effectiveness of the PowerBERT-based framework in detecting and identifying two common types cyber attacks in AGC in power grids, and it has a better performance in the downstream cyber attacks classification than other signal processing based and DL/ML based models. Our proposed loss function, SME, can effectively address the common data imbalance problem encountered in real-world applications. We believe that our framework can be readily applied to other scenarios, as it only requires easily accessible unlabeled data and a small amount of labeled data to achieve superior performance. Furthermore, the proposed loss function can be widely employed in other unsupervised learning models to manage the data imbalance issue.
%\section*{Acknowledgement}
%This research is supported in part by the National Research Foundation, Singapore, and the Energy Market Authority, under its Energy Programme (EP Award No. NRF2018-NCR003-0013), and in part by the National Research Foundation, Prime Minister’s Office, Singapore under its Campus for Research Excellence and Technological Enterprise (CREATE) programme.
% \clearpage
\printbibliography
% \bibliographystyle{plain}
% \cite{Liu1}
% \cite{Xu2}
% \cite{He3}
% \cite{Devlin4}
% \cite{Schneider5}
% \cite{Li6}
% \cite{Yang7}
% \cite{Inoue8}
% \cite{Ganesh9}
% \cite{Yuan10}
% \cite{Defu11}
% \cite{louxin12}
% \cite{Siegel13}
% \cite{kundur14}
% \cite{liu15}
% \cite{Tan16}
% \bibliography{ref}
\end{document}